\documentclass[letterpaper, 10 pt, journal, twoside]{ieeetran}  

\usepackage{amssymb,graphicx,amsmath,color,algorithm,algorithmic,url}
\usepackage[english]{babel}
\usepackage{blindtext}
\usepackage{gensymb}
\usepackage{comment}
\usepackage{makecell}
\usepackage{afterpage}
\usepackage[noadjust]{cite} 
\usepackage{soul}
\usepackage{hhline,caption}
\usepackage{hyperref}
\usepackage[font=small,labelfont=bf]{caption}

\IEEEoverridecommandlockouts                              

\title{TacMMs: Tactile Mobile Manipulators for Warehouse Automation}

\author{Zhuochao He, Xuyang Zhang, Simon Jones, Sabine Hauert, Dandan Zhang, Nathan F. Lepora$^{1}$
\thanks{Manuscript received December 22, 2022; Revised April 3, 2023; Accepted May 19, 2023.
This paper was recommended for publication by Editor Ashis Banerjee upon evaluation of the Associate Editor and Reviewers' comments.
This work was supported by an award from the Leverhulme Trust on ‘A biomimetic forebrain for robot touch’ (RL-2016-39).
{\em (Corresponding author: Nathan F. Lepora)}}
\thanks{$^{1}$The authors are with the Department of Engineering Mathematics and Bristol Robotics Laboratory, University of Bristol, Bristol BS8 1UB, U.K. (email: n.lepora@bristol.ac.uk).}%
\thanks{Digital Object Identifier (DOI): see top of this page.}
}
\begin{document}
\maketitle

\begin{abstract}
Multi-robot platforms are playing an increasingly important role in {\color{black}warehouse automation for efficient goods transport}. This paper proposes {\color{black}a novel customization of a multi-robot system, called} Tactile Mobile Manipulators (TacMMs). Each TacMM integrates a soft optical tactile sensor and a mobile robot with a load-lifting mechanism, enabling cooperative transportation in tasks requiring coordinated physical interaction. More specifically, we mount the TacTip (biomimetic optical tactile sensor) on the Distributed Organisation and Transport System (DOTS) mobile robot. The tactile information then helps the mobile robots adjust the relative robot-object pose, thereby increasing the efficiency of load-lifting tasks. This study compares the performance of {\color{black}using two TacMMs} with tactile perception with traditional vision-based pose adjustment for load-lifting. The results show that the average success rate of the TacMMs (66$\%$) is improved over a purely visual-based method (34$\%$), with a larger improvement when the mass of the load was non-uniformly distributed. {\color{black}Although this initial study considers two TacMMs, we expect the benefits of tactile perception to extend to multiple mobile robots.} Website: \mbox{\url{https://sites.google.com/view/tacmms}.}
\end{abstract}

\begin{IEEEkeywords} Tactile Sensing, Multi-robot system, Warehouse transportation \end{IEEEkeywords}

\section{INTRODUCTION}

With the increasing number of online customized orders, there are higher requirements for transportation and warehouse management \cite{custodio2020flexible}. Currently, traditional mobile robots, such as Automated Guided Vehicles (AGVs) \cite{vis2006survey} and forklifts \cite{wang2010innovative}, automatically localize products then lift and transport them in the warehouse. Of these, load-lifting is an important step for warehouse automation, for which a robot must: 1) perceive the pose of the load; 2) determine the optimal lifting position on the target based on the perceived pose; and 3)~control the contact pressure to achieve robust lifting and avoid damage to the load. However, due to difficult to precisely control object contact for robust lifting, traditional visually-guided mobile robots are limited for transporting goods with a distal non-contact modality. 


\begin{figure}[t!]
	\centering
	\begin{tabular}[b]{@{}c@{}}
        \includegraphics[width=\columnwidth,trim={0 0 0 0},clip]{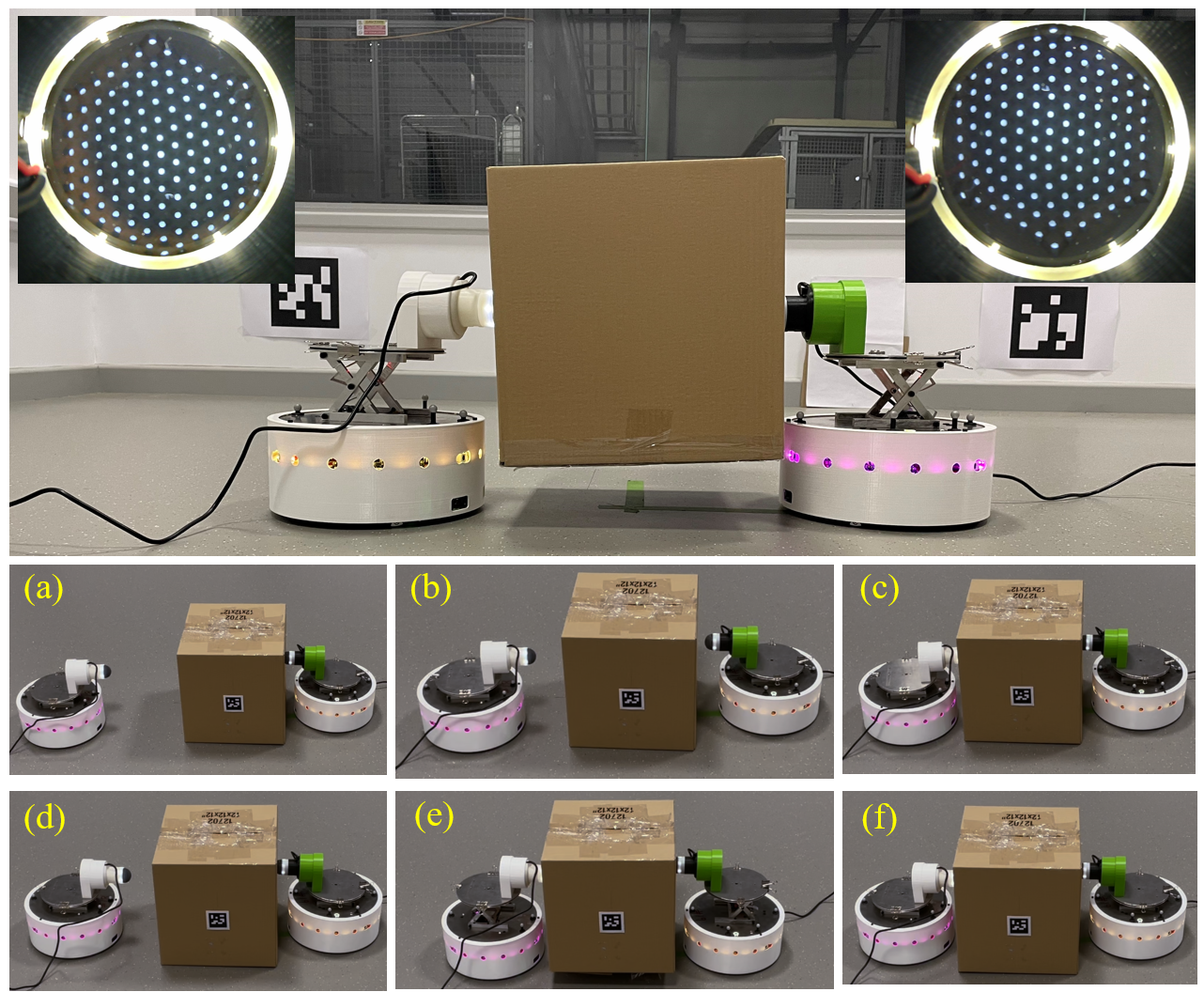}
    \end{tabular}
	\caption{\textcolor{black}{TacMM system lifting a box. Top: a box lifted by two DOTS mobile robots each with a TacTip {\color{black}optical tactile sensor} mounted on a raisable platform; tactile images also~shown. Bottom: steps to lift a box: (a) approach object; (b) adjust pose after initial contact; (c) establish a second contact with object; (d) adjust pose after the second contact; (e) lift object; (f) lower object.}}
	\label{fig:1}
\end{figure} 

Motivated by these shortcomings, multi-robot cooperative systems with soft end-effectors have been developed for warehouse transportation~\cite{schou2018skill,liu2014dynamic,basile2012task}, resulting in more efficient and robust systems, with higher performance and the ability to lift heavier products. The transportation strategies of this kind of robotic system include pushing, grasping and caging \cite{tuci2018cooperative,adouane2004hybrid,khatib1999robots}. Moreover, grasping or lifting with multiple robots is analogous to using the fingers of a human or robot hand, where it is known that to achieve safe lifting the fingers must pre-adjust to a desired pose and apply a reasonable force. In multi-robot cooperative systems, the robots commonly rely on an external vision system to feedback the relative pose of the target objects. However, as is well known in robot grasping, such vision systems are limited by occlusion, calibration issues and inaccuracy from a camera situated away from the target~\cite{matak2022}. 

In contrast, tactile sensing offers the capability to estimate the relative robot-object pose for the lifting task without the aforementioned issues of vision. Here we use an optical tactile sensor called the {\color{black}TacTip (Tactile fingerTip) \cite{ward2018tactip,lepora2021soft} which has a 3D-printed soft dome-like structure mounted over an internal camera and lighting. This sensor is well-suited for Tactile Mobile Manipulators (TacMMs), being of the right size and shape to mount on the top liftable platform of DOTS (Distributed Organization and Transport System) mobile robots designed for cooperative automation~\cite{jones2022dots}. Furthermore, both the TacTip and DOTS are open-source and easily fabricated, enabling others to customize and build upon this work.} 


To the best of our knowledge, the TacMM represents the first multi-robot system that integrates a high-resolution soft tactile sensor into mobile manipulators for warehouse automation. The main contributions of this work are as follows:\\
\textcolor{black}{\noindent 1) We propose a novel tactile multi-robot system, which integrates the TacTip and DOTS mobile robots, with application to warehouse transportation and logistics (Figures 1,2).}\\
{\color{black}\noindent 2) We show the tactile sensors are effective at estimating the pose of the contact surface, and propose a tactile servo control policy to adjust the robot to a desired pose on the object.\\
\noindent 3) We successfully demonstrate the load-lifting task with two TacMMs using the tactile feedback to work together collaboratively to improve the stability of lifting the load.\\
\noindent In this work, we introduce the tactile mobile manipulator concept with a minimal configuration of two TacMMs; however, we expect this paradigm for mobile manipulation will be far more effective with multiple tactile manipulators working collaboratively to handle large and complex loads.}


This paper is organized as follows. First, we review warehouse robots and their control systems. The hardware, software and vision/tactile movement strategies for the TacMM system are described, and experiments are illustrated. Experiments are conducted to estimate the relative pose estimation errors and compare the performance of TacMM with the baseline (vision only) system. Finally, we summarise the experiment results and discuss the future work and limitations of this system.

\section{RELATED WORK}

\subsection{Lifting-based transport strategies}
Transporting objects by lifting can avoid frictional damage of objects caused by traditional pushing strategies if there is no carrier underneath the object \cite{hichri2019design}. In \cite{hirata2000coordinated} and \cite{kume2002decentralized}, Kume et al proposed a virtual 3D caster control in the leader-follower decentralised system. Based on the virtual 3D caster, the follower robots were able to estimate the motion of the leader. In \cite{hichri2016cooperative}, Hichri et al applied Force Closure Grasping (FCG) to determine the position of each robot before they try to lift the object. After lifting, the object is placed on top of the robot. However, one of the requirements of FCG is that the robot must adjust its pose until its end-effector is normal to the contact surface of the target before lifting. 

\subsection{Visual pose estimation}
Inspired by human lifting behavior, pose estimation and adjustment can be completed before lifting the object, resulting in a more efficient and stable lifting behavior. Currently, transportation robots still rely on vision to detect the pose of the target object, such as \cite{du2021vision} detects the 6D pose of the target for manipulation. Also using 6D pose, a robot arm is controlled to approach the estimated pose to guide a gripper to the target \cite{yang2012monocular}. In warehouse transportation, vision systems can give an approximate pose of the target, but do not provide contact information needed for object manipulation. 

Another solution is to integrate the vision system into the mobile manipulator to provide local information such as the relative pose and shape of the target object \cite{guangrui2017vision}. Usually, these local systems use stereo vision, which has a minimum usable camera line-of-sight that is not suited for nearby objects. Moreover, local vision cannot provide contact information required for stable lifting, particularly for delicate objects. Because of these drawbacks, we focus here on a system fusing visual and tactile feedback or just using tactile alone.

\begin{figure}[t!]
	\centering
	\begin{tabular}[b]{@{}c@{}}
        \includegraphics[width=\columnwidth,trim={0 0 0 0},clip]{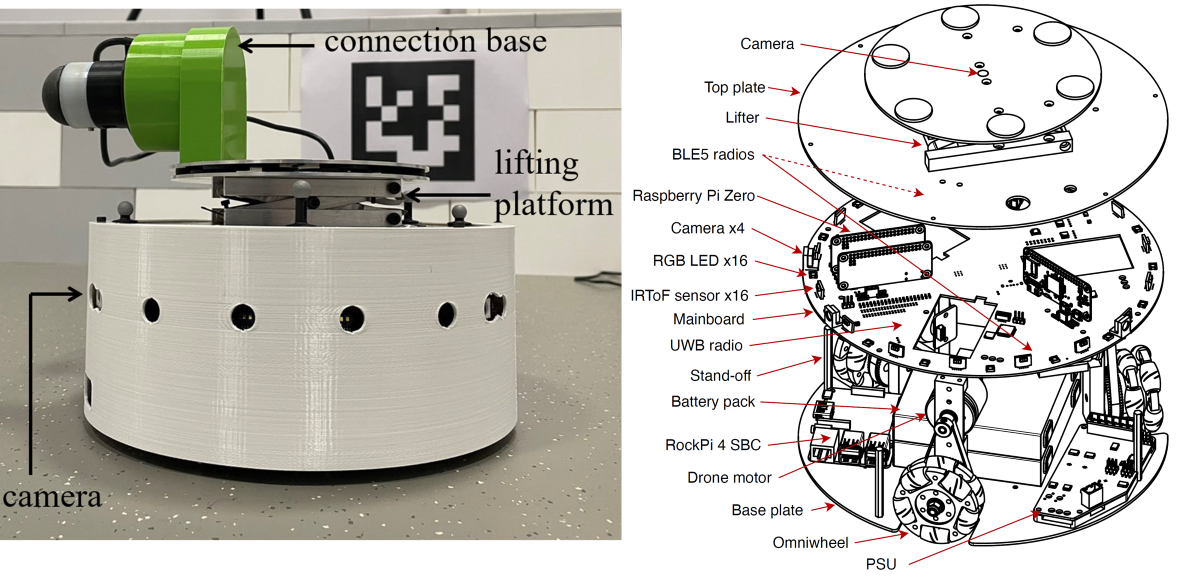} 
    \end{tabular}
	\caption{\textcolor{black}{Overview of the TacMM tactile mobile manipulator. Left: the DOTS distributed organization and transport system mounted with a TacTip soft high-resolution tactile sensor on its lifting platform. Right: schematic of the DOTS mobile robot, with two of the four cameras in the base visible to the left.}}
	\label{fig:2}
\end{figure} 

\begin{figure}[t!]
	\centering
	\begin{tabular}[b]{@{}c@{}}
        \includegraphics[width=\columnwidth,trim={75 0 75 0},clip]{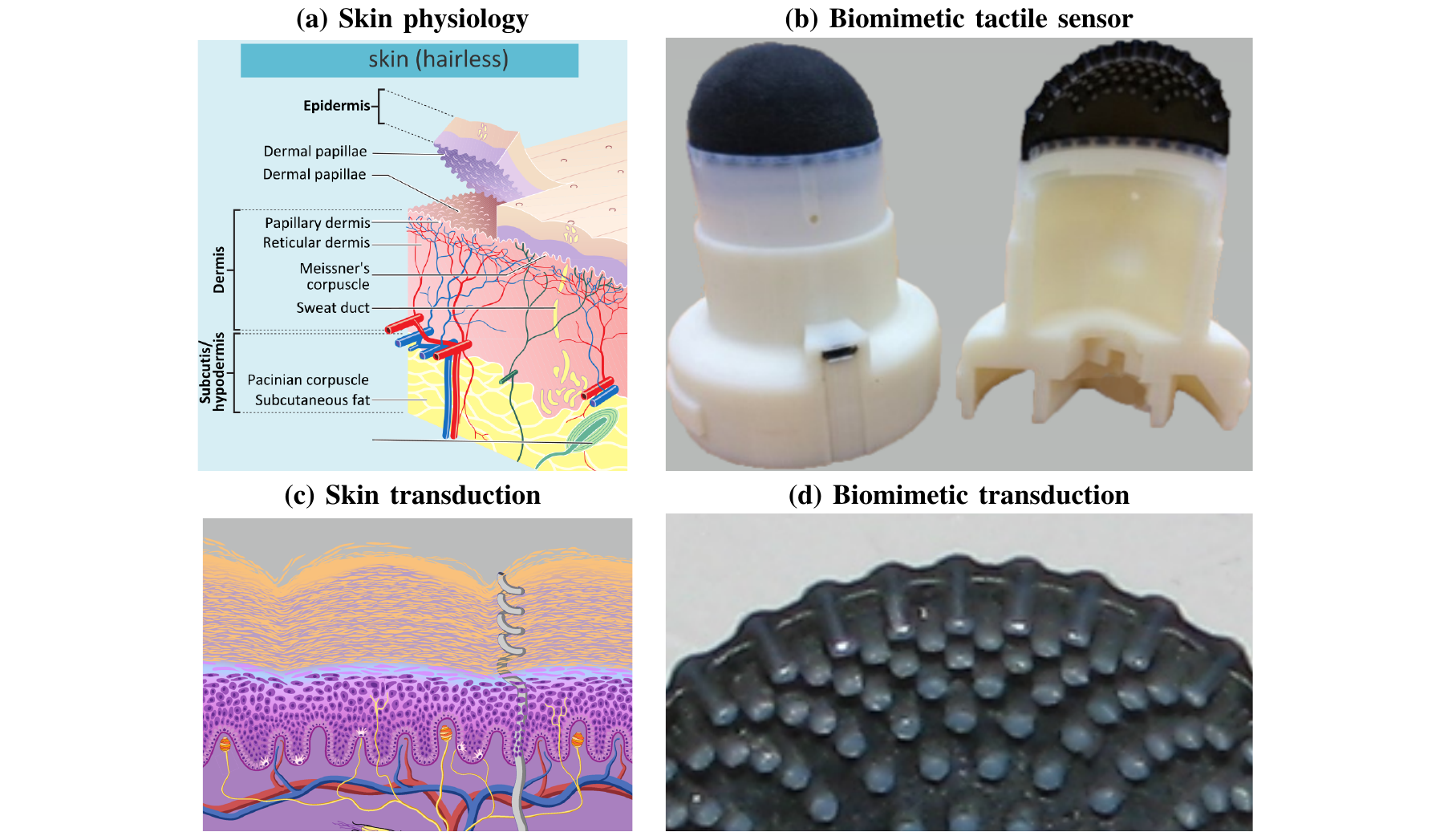} 
    \end{tabular}
	\caption{\textcolor{black}{Overview of the TacTip tactile sensor concept and construction. (a,b) skin physiology and transduction via internal dermal papillae where the mechanoreceptors are located. (c,d) multi-material 3D-printed structure of the TacTip featuring biomimetic papillae tipped with markers that are imaged with an internal camera (images from Ref.~\cite{lepora2021soft}).}}
 \label{fig:2a}
\end{figure} 

\subsection{Tactile pose estimation}

{\color{black}Pose estimation using tactile sensing has been studies in detail because it is needed to controlling robot hands and other end effectors using tactile feedback, and underlies methods for tactile servoing to control contact with an unknown object~\cite{li2013, lepora2021pose}. High-resolution tactile sensing using an internal camera to image deformation of the sensing surface offers the opportunity to leverage advances in computer vision with convolutional neural networks, which has been applied to accurate estimation of surface hardness~\cite{yuan2017}, shape~\cite{suresh2022} and pose~\cite{lepora2019} amongst others. In this study, we use a high-resolution, optical tactile sensor (the TacTip, Fig.~3) that has established capabilites for pose estimation and tactile servoing~\cite{lepora2020optimal,lepora2021pose,lepora2021soft}. 

A subtlety with pose estimation for soft tactile sensors is that the manner of contact affects the deformation of the tactile sensor along with the pose of the contacted surface~\cite{lepora2019,lepora2020optimal}; for example, shearing from the left or right to a given pose will produce different tactile images. Therefore, for effective tactile servo control, the neural network needs to be trained to ignore the effects of shear upon contact, which can be done by introducing random shear perturbation during the data collection~\cite{lepora2020optimal,lepora2021pose}. This leverages that deep neural networks are highly effective at predicting labelled quantities and ignoring unlabelled variations on complex data. As a consequence, tactile pose estimation with the TacTip has been successfully applied to range of tasks including object exploration and contour following~\cite{lepora2019,lepora2021pose}, non-prehensile manipulation and object pushing~\cite{lloyd2021}.}


\section{METHODOLOGY}
\label{sec:3}

\subsection{Hardware methods}

The proposed TacMM tactile mobile manipulator system comprises three parts: (i) the DOTS distributed organization and transport system mobile robot; (ii) the mounted TacTip tactile sensor; and (iii) a custom 3D-printed connecting base (see Fig.~\ref{fig:2}).

{\color{black}\subsubsection{DOTS mobile robot system} this mobile robot features a wheeled omnidirectional base with 4 cameras mounted in the base and an actuated lifting platform (Fig~2).} The overall system includes an integrated remote development platform and physical mobile robots \cite{jones2022dots}, which allows researchers to develop this system in simulation before physical experimentation. The integrated remote development platform is based on Robot Operating System version 2 (ROS2) and utilizes Gazebo. This robot has omniwheels and its lifting platform is capable of a 2\,kg maximum payload. The multiple cameras provide a local vision system with stereo at the front and additional rear/side viewpoints.

{\color{black}\subsubsection{Tactile end effector} The high-resolution tactile fingertip (TacTip) is mounted using a connection base on top and oriented to the front of the mobile robot (Fig.~\ref{fig:2}). The standard TacTip is about 100\,mm long with a 40\,mm diameter hemispherical soft dome, which is well-suited for mounting horiontally on the lifting platform to protrude to the front. Then the lifting platform gains a new functionality as being a vertical actuator for a soft tactile fingertip.

The TacTip is a camera-based tactile sensor with a biomimetic 3D-printed soft skin based on the physiological structure of human skin~\cite{lepora2021soft} (Fig.~\ref{fig:2a}. A TacTip is composed of a built-in camera, a mounting base, a LED ring for illuminating the internal skin, and a 3D-printed skin with internal pins that mimic the structure of the dermal-epidermal boundary~\cite{ward2018tactip,lepora2021soft}. The tips are designed to be modular and replaceable, with an acrylic window below the 3D-printed biomimetic skin, between which a gel is injected to give the tip a softness close to adipose tissue of the human fingertip. When the tip of the TacTip contacts an object, the skin deformation causes the internal pins to lever which amplifies the sideways motion of the markers captured by the internal camera.

For applications such as contact pose estimation, using the tactile image as the input to a convolutional neural network is the easiest and most robust method~\cite{lepora2021soft} (compared to extracting and processing the marker locations, for example.) Therefore, OpenCV is used capture and pre-process the tactile images to view cropped view of just the markers, convert the image to grey-scale then binarize with an adaptive threshold, then sub-sample to 128$\times$128 pixels for efficient learning and prediction. These pre-processed tactile images are fed into a PoseNet, a pose estimation neural network \cite{lepora2020optimal} to predict the relative sensor-object contact pose, including contact orientation and depth, whose training is described later.}



\subsection{System design and control}
\label{sec:3a}

The effectiveness of the TacMM system is first evaluated in simulation before real-world deployment. 

\begin{figure}[t!]
	\centering
	\begin{tabular}[b]{@{}c@{}}
        \includegraphics[width=0.9\columnwidth,trim={0 0 0 0},clip]{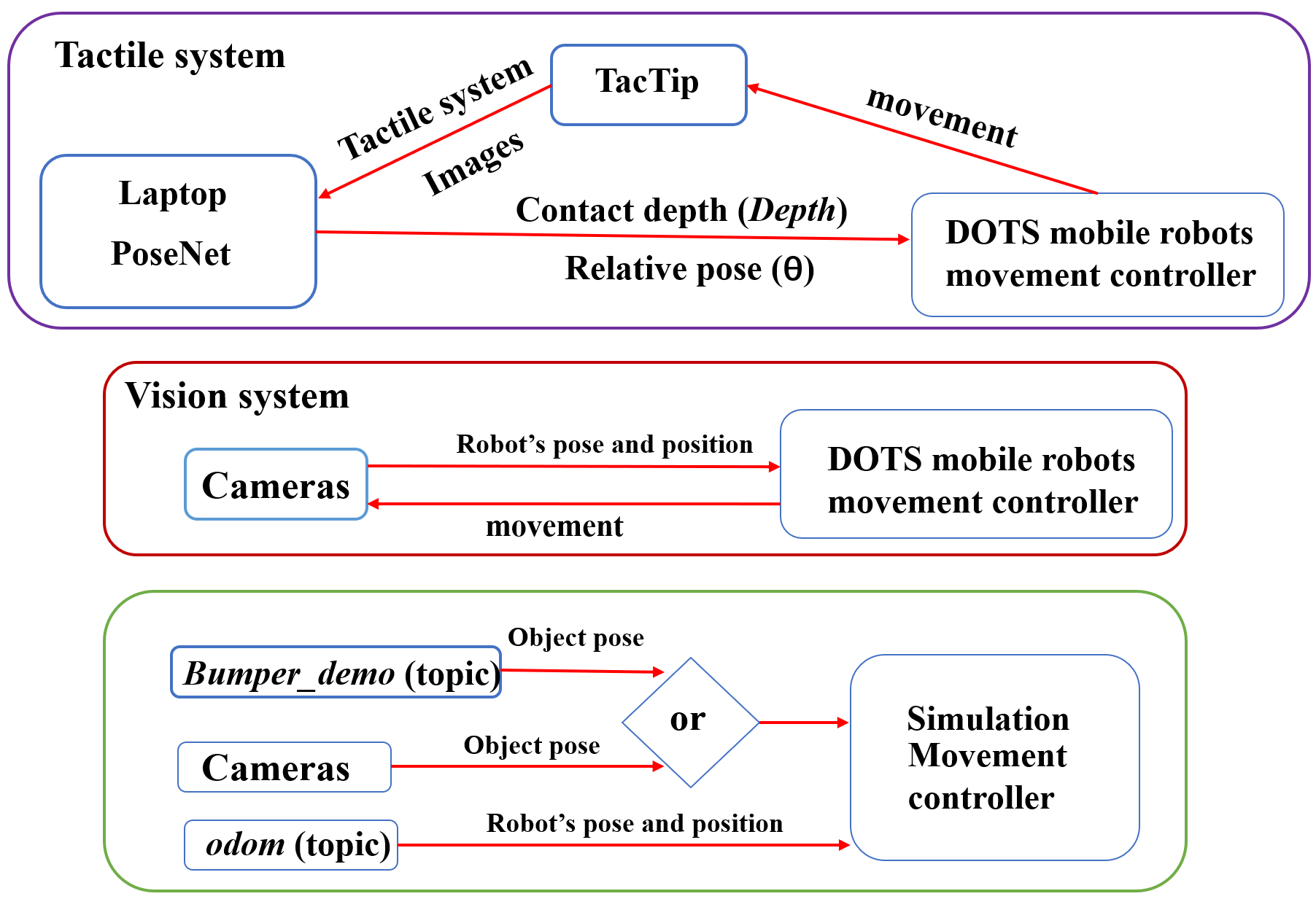} 
    \end{tabular}
	\caption{\textcolor{black}{Tactile and vision control system architectures.}}
	\label{fig:3}
\end{figure}

The control methods for each mobile robot are based on ROS2, which processes the sensor data, controls the robot motion and implements communications between robots (Fig.~\ref{fig:3}). The tactile-based system adjusts the relative pose between the mobile robot and the object sequentially over multiple contacts. The vision system is an open-loop control system that adjusts the motion planning before contact, with no feedback during the contact and the lifting process. 

\subsubsection{Simulated system}
{\color{black}For validation, we also use simulated versions of the real vision and tactile sensors described below.} In the simulation, all experiments rely on ROS2. For example, the camera is a plug-in from Gazebo and the tactile sensors that will be used to estimate the relative pose of the target are substituted by a bumper {\color{black}that returns the exact pose on contacting the object. This allows us to both assess the effectiveness of the TacMMs in simulation alongside the real-world deployment.}

{\color{black}\subsubsection{Real vision system}
An external camera is mounted on a tripod with a view of the robot and TacTip from above, with the relative pose of the TacTip and object} detected using ArUco markers. The robot then adjusts its pose until the central line of the TacTip end effector is normal to the surface of the target object. The TacTip deformation upon contact is captured by an internal camera and the tactile images are used as an input into the PoseNet neural network \cite{lepora2020optimal}. 

\begin{figure}[t!]
	\centering
	\begin{tabular}[b]{@{}c@{}}
        \includegraphics[width=0.85\columnwidth,trim={0 145 0 15},clip]{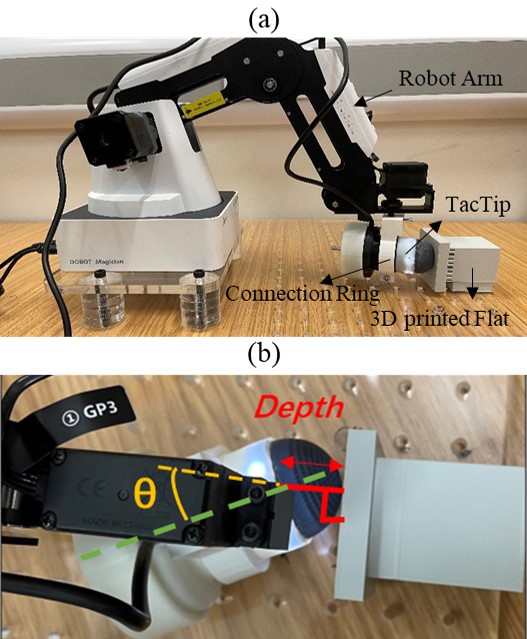} \\
        \includegraphics[width=0.85\columnwidth,trim={0 0 0 180},clip]{Fig4a.jpg} 
    \end{tabular}
	\caption{\textcolor{black}{Top: A dobot Magician is used to gather pose-labelled training and test tactile images for training, validation and testing the pose-prediction network. Bottom: Labelled angle and depth parameters used in the data collection.}}
	\label{fig:4}
\end{figure}

{\color{black}\subsubsection{Real tactile system}
The TacTip  is mounted as described in the previous hardware section and used to collect tactile images of the contact against the object. The contact depth \textit{Depth} and the relative angle $\theta$ between the normal of the object and the TacTip (Fig.~\ref{fig:4}), using a PoseNet neural network whose training and architecture will be described later.} In addition, the contact \textit{Depth} prediction will also be used to distinguish between contact and non-contact situations. We aim for a control strategy that tunes the \textit{Depth} until it reaches a maximum threshold, then adjusts the angle $\theta$ of the TacTip. 


\begin{figure}[t!]
	\centering
	\begin{tabular}[b]{@{}c@{}}
	    {\bf Control strategies used for pose adjustment}\\
        \includegraphics[width=0.9\columnwidth,trim={0 0 0 0},clip]{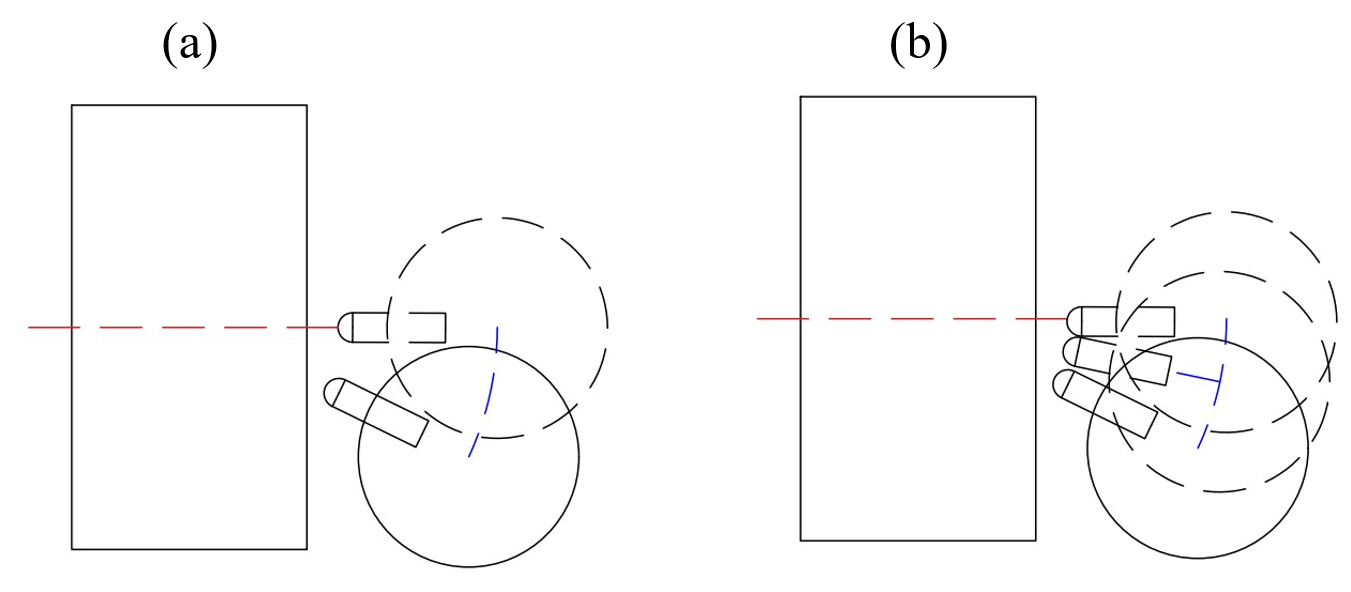}\\ 
	    {\bf The examples of the FCG}\\
        \includegraphics[width=0.85\columnwidth,trim={0 0 0 0},clip]{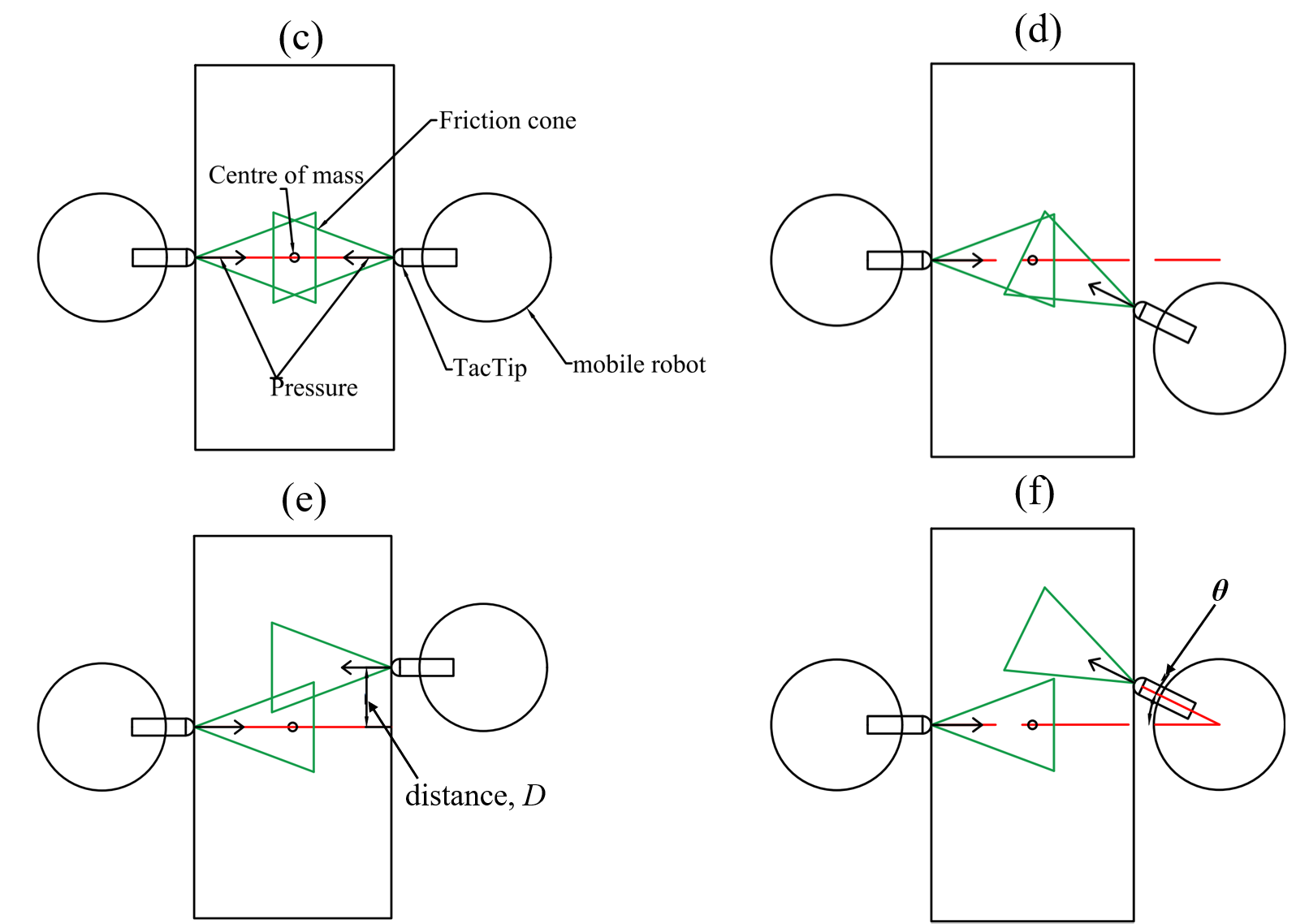}\\ 
    \end{tabular}
	\caption{\textcolor{black}{Strategies where (a) The robot rotates around its centre, (b) the robot rotates around an object's centre and (c) the robot rotates around an object's centre contacting it several times. (d) and (e) illustrate satisfying the FCG, while (f) and (g) illustrate failure cases according to the friction cones.}}
	\label{fig:5}
\end{figure}

{\color{black}\subsubsection{Multi-robot system} After} adjusting their poses, the two TacMMs must cooperate to lift the object. The robot poses should satisfy the Force Closure Grasp (FCG) condition from \cite{hichri2016cooperative}, assuming that the force from each TacTip is perpendicular to the object surface. Specifically, the forces on the object form two friction cones (Fig.~\ref{fig:5}) that, to lift the object, the summed forces and summed torques must equal zero. This also assumes the centre of mass of the object is within the intersection of all friction cones. The examples in Fig.~\ref{fig:5}(d)-(g) demonstrate the requirements of the FCG for two robots, leading to constraints:\\
\textcolor{black}{\noindent I) The two robots must be on opposite sides of the object.}\\
\noindent II) The contact position of the TacTip must be as close as possible to the centre of the contacted flat surface of the object. \\
\noindent III) The angle between the normal of the contacted flat surface and the contact pose should be as small as possible.\\
\noindent IV) The contact depths are equal and the summed torques should be zero. 

{\color{black}\subsubsection{Control strategies via behaviour trees}
Two }tactile control strategies are proposed for pose adjustment (Fig.~\ref{fig:5}(a,b)), both of which rrotate around the object's centre (estimated using the vision system), with the first using a single contact (a) and the other using multiple contacts (b). Similar control strategies will also be used to assess visual control. All strategies use Behavior Trees (BTs) for robot motion control \cite{marzinotto2014towards}. Every movement behavior is considered as a branch in each strategy within the behavior tree. The attribute of every branching contains \textit{Sequence}, \textit{Selector} and \textit{Parallel} operations. Furthermore, the strategies for lifting an object in the tactile and visual systems are also in the form of BTs.

The BT for the self-rotation controller shown in Fig.~\ref{fig:5}(a) is similar to the object-rotation controller Fig.~\ref{fig:5}(b), each having three parts comprising \textit{TacTip\_information}, \textit{Gather} and \textit{Action\_order}. \textit{TacTip\_information} receives the pose from the PoseNet for the physical experiments (with a bumper used instead in simulation). \textit{Gather} sets up the ROS2 subscription and publisher in the strategy. \textit{Action\_order} commands the robot towards the object until the bumper or TacTip contact, then to move back and adjust its pose according to the sensor feedback; this is repeated until the contact depth reaches a threshold for the lifting task. In the Fig.~5(a), after the bumper or the TacTip returns the results of the relative pose, the mobile robot will rotate based on a proportional gain until the orientation of the TacTip is normal to the target. In the BT in Fig.~5(b), the robots rotate around the centre of the object which is assumed known in advance.

{\bf TacMM pose adjustment strategy:} This strategy contacts the target multiple times (Fig.~\ref{fig:5}(c)). The BT repeats \textit{Action\_order} to contacting the object multiple times, stepping back and rotating around the object centre (estimated from the vision system) until the absolute value of the predicted angle is less than a preset angle threshold.


{{\bf Vision-based pose adjustment strategy:}} The BT of this strategy is similar to the TacMM pose adjustment strategy, instead using an ArUco marker for pose estimation between the TacMM and the target object. 

{{\bf TacMM lifting strategy:}} This strategy uses the optimal tactile strategy (see Section \ref{sec:4b} later) to lift an object, with each robot in the lifting task adjusts its pose as in the tactile pose adjustment strategy. After each robot completes its pose adjustment, this strategy ensures they communicate with each other, allowing both robots to concurrently activate the lifting platform and then lower it (example shown Fig.~\ref{fig:1}).

{{\bf Vision-based lifting strategy:}} This strategy is the Vision-based pose adjustment strategy counterpart of the TacMM lifting strategy, giving a baseline in the load-lifting task. After adjusting the robots' pose according to vision, the robots communicate with each other to concurrently lift and lower the platform.

\subsection{Experiment details and model training}
\label{sec:3c}

Five experiments were conducted. Training and testing of the tactile pose estimation was conducted both offline on a test set and online. The third experiment was in \textit{Gazebo} simulation and the last two experiments compared the performance of TacMM with traditional vision-based control as a baseline. 

{\color{black}\subsubsection{{\bf Tactile pose model and testing}} 
The first experiment focused on building a model for pose estimation, which we call PoseNet. We needed to gather training data of tactile images with labelled poses, for which we used a Dobot Magician robot arm with the TacTip mounted as end effector contacting a 3D-printed flat surface (Fig.~\ref{fig:4}). The Dobot Magician is a low-cost four-axis desktop robot arm (Fig.~\ref{fig:4}(d)) accurate to 0.2\,mm. The TacTip is detached from the robot manipulator and mounted on the robot arm} with a 3D-printed connection ring (Fig.~\ref{fig:4}(b)). A laptop (AMD Ryzen 7 5800H with NVIDIA GeForce RTX 3060 GPU) is used to set the robot arm end effector pose, store tactile images and train/test the PoseNet.  


During data collection, the robot arm moved the TacTip to contact the 3D-printed flat surface at random depths and angles within a set range, and recorded the corresponding poses and depth as labels. 5000 tactile images were collected with pose angles $\theta\in[-25^\circ,25^\circ]$ and contact depths $\in [1,5]\,$mm. {\color{black}These tactile images were captured after a random horizontal linear and rotation shear of less than 5\,mm and 5 degrees, which as discussed in the background is necessary for robustness to how the sensor contacts the object~\cite{lepora2020optimal}. An addition set of 500tactile images were gathered at a non-contact depth, where the depth label was set to $0\,$mm, enabling the PoseNet to serve as a predictor of contact/non-contact.} 

For PoseNet model training and testing, the 5500 tactile images were randomly split into a 75$\%$ for training and a 25$\%$ for testing. We refer to \cite{lepora2020optimal} for the hyperparameters and other implementation details. The Mean Absolute Error (MAE) of the \textit{Depth} and $\theta$ are used to measure performance.

\subsubsection{{\bf Online tactile pose prediction testing in real environment}} 

The trained tactile pose prediction PoseNet model was verified in the physical environment, by controlling the TacTip to interact with the 3D-printed flat test surface at a known relative contact depth and angle. The trained PoseNet predicted the angle and depth, repeating 5-fold every 5° from -25° to 25° and every 1\,mm from 1 mm to 5 mm. The MAE of the relative angle and contact depth and the PoseNet computation time are used as metrics.

\subsubsection{{\bf Tactile pose adjustment in simulation}} 

The performance of pose adjustment for TacMM was evaluated in simulation, by substituting the TacTip with a bumper that returns the precise pose of the contact object. This experiment assessed errors from the kinematics of the DOTS mobile robot in simulation.

The Gazebo simulation was used to test the performance of the pose adjustment. A cube of size 60\,mm$\times$30\,mm$\times$30\,mm and variable orientation was located 1.1\,m in front of the mobile robot. For every 5° from -25° to 25°, the robot repeated the pose adjustment using the single and multiple-contact pose adjustment strategies (Figs~\ref{fig:5}(a,b)) repeated 5-fold each. We measured the angle error $\theta$ and the translation distance \textit{D} along the contact surface, measured between its centre and the contact point (Figs~\ref{fig:5}(f,g)). The angle and distance MAEs at various cube poses are used as performance metrics.

\subsubsection{{\bf Tactile and visual pose Adjustment in real environment}} 

The TacMM and visual control systems were then compared using the pose adjustment with the real robot. A plastic cube (weight 8.58\,kg) was used in the TacMM experiment, and replaced by a box with ArUco marker in the vision experiment. A tripod-mounted camera videoed the pose adjustment from a top view.

We conducted 55 experiments with the TacMM and vision systems, with the same experiment settings as in the simulated pose adjustment above. The TacMM and vision systems used their respective pose adjustment strategies (1 and 2) described above. The distance error was measured by a ruler and the angle error by OpenCV detection of the ArUCo marker. The angle and distance MAEs are used as performance metrics.

\subsubsection{{\bf Load Lifting}} 

The performance of the load lifting was tested using the TacMM and vision systems. Two robots were controlled identically, with a box placed between them. Two ArUco markers on the box were used for tactile validation and for the vision-based strategy.

Varied attributes of the box included: 1) empty box, 2) two empty boxes stacked vertically, 3) a 200g weight on the top of an empty box, 4) a 200g weight on the bottom of an empty box and 5) a payload weighted almost 500g on the top of an empty box (Fig.~\ref{fig:8}(a)). For each attribute, the experiment was tested 10 times and success rate recorded.

\section{RESULTS AND ANALYSIS}

\subsection{Evaluation of tactile pose prediction performances}

The performance of tactile pose prediction with the PoseNet is first described in terms of the pose adjustment and object lifting tasks. The performance of the tactile pose prediction is then quantified in an online experiment, and a threshold is established to distinguish between contact and non-contact states.

\begin{figure}[t!]
	\centering
	\begin{tabular}[b]{@{}c@{}}
		\includegraphics[width=\columnwidth,trim={0 0 0 0},clip]{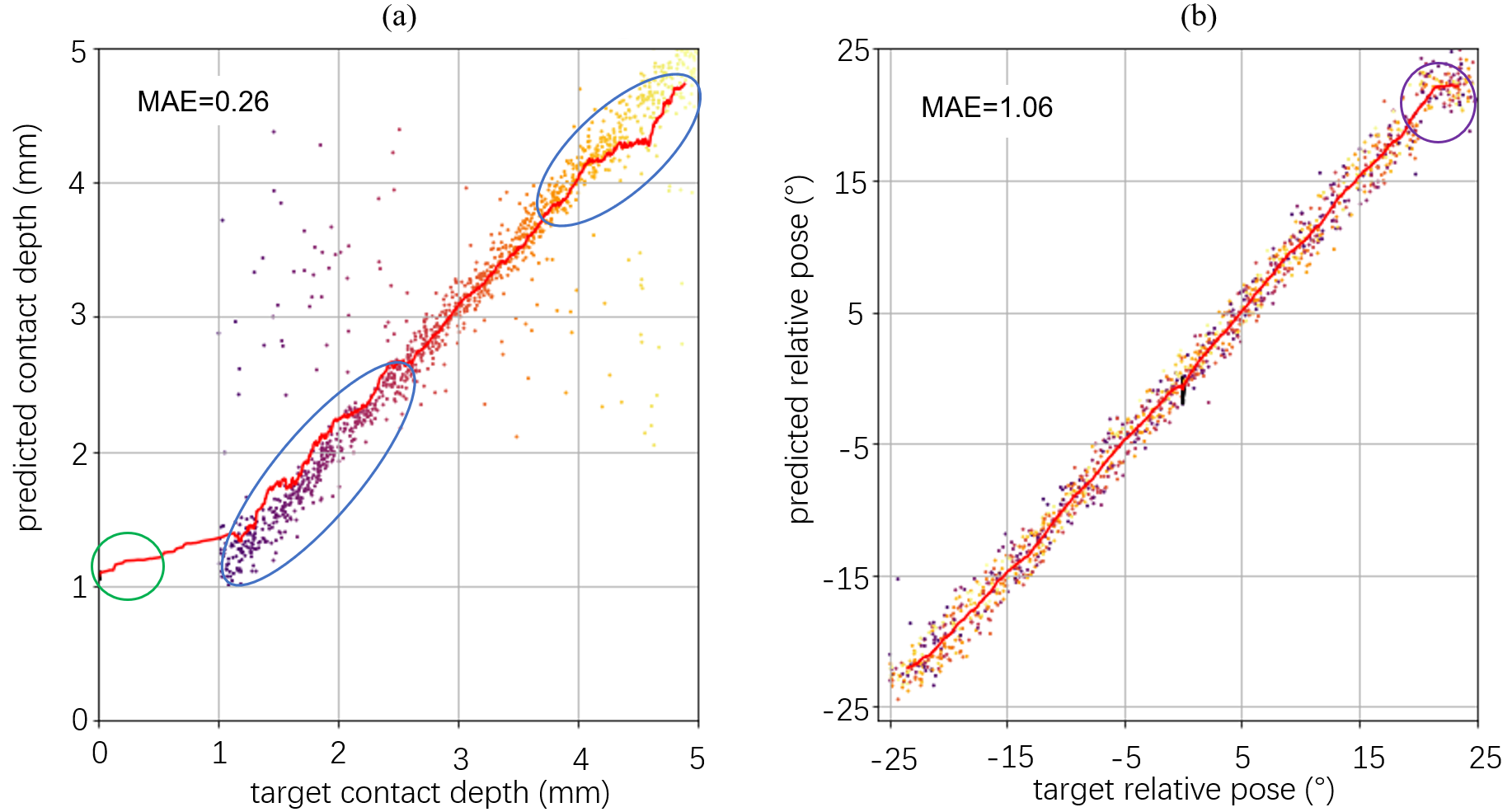}
	\end{tabular}
	\caption{Pose prediction performance for (a) contact depth and (b) angle, using a robot arm. Performance was good with the elliptical regions indicating regions of interest, demonstrating the depth network can distinguish between non-contact (labelled 0\,mm) and contact data (labelled 1\,mm and above).}
	\vspace{0em}
	\label{fig:6}
\end{figure}

Overall, the MAE of the contact depth and the angle are 0.26\,mm and 1.06° respectively (Fig.~\ref{fig:6}), when using a robot arm to collect a dataset containing both contact and non-contact tactile images. We judge this performance of the trained model as sufficiently accurate for the pose adjustment task and the object lifting task. That said, it is worht noting that the accuracy of the contact depth in the range from 4\,mm to 5\,mm and from 2.5\,mm to 1\,mm (blue ellipse in Fig.~\ref{fig:6}) is less that that of the 2.5\,mm to 4\,mm range. Further, when the angle is over 20°, the accuracy of the model declines slightly (Fig.~\ref{fig:6}). 

Although the depth of a non-contact tactile image is labelled as 0\,mm, the predicted value has a discrepancy in being close to 1\,mm. Thus, there is a question whether the non-contact images during training have influenced the predictions at other contact depths. To examine this, we used a dataset composed solely of tactile images from the contact situation to train and test the PoseNet. The error of predicting contact depth improved slightly to 0.19\,mm and the angle predictions were little affected (Table~\ref{tab:1}). Given these relatively small changes, we will train with non-contact images so the PoseNet can distinguish non-contact and sufficiently contacting states. 

\begin{table}[t]
    \centering
	\caption{Pose prediction performance trained with and without non-contact tactile images collected using a robot arm.}
	\vspace{0em}
	\begin{tabular}{@{}c|cc@{}}
    	\textbf{Dataset} & \textbf{MAE of \textit{Depth} (mm)} & \textbf{MAE of $\theta$ (°)}\\
    	\hline\hline
    	With non-contact images & 0.26 & 1.06 \\
    	 \hline
    	Without non-contact images  & 0.19 & 1.07 \\
    	 \hline
    	\end{tabular}
	\label{tab:1}
	\vspace{0em}
\end{table}

\begin{figure}[t!]
	\centering
    \begin{tabular}[b]{@{}c@{}c@{}}
        \textbf{\footnotesize{(a)}} & \textbf{\footnotesize{(b)}}  \\
	    \includegraphics[width=0.49\columnwidth,trim={0 0 0 0 },clip]{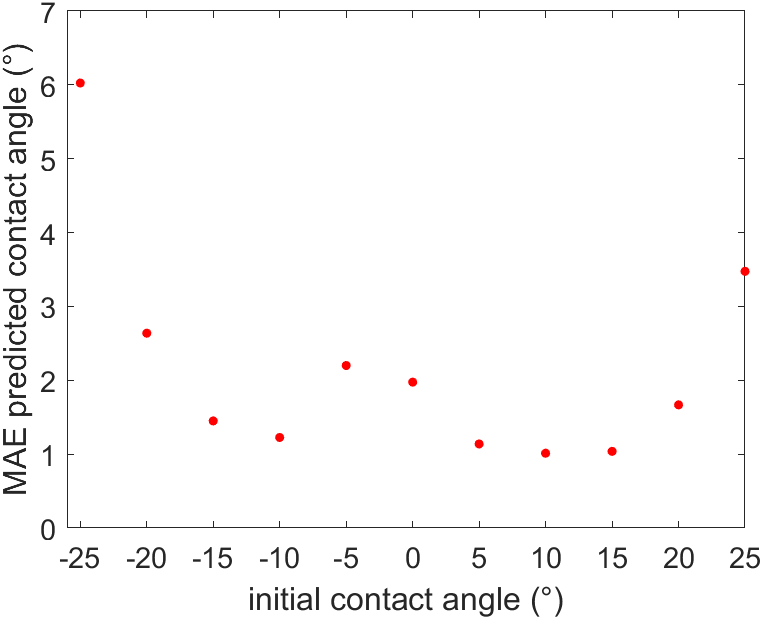} & 		
		\includegraphics[width=0.49\columnwidth,trim={0 0 0 0},clip]{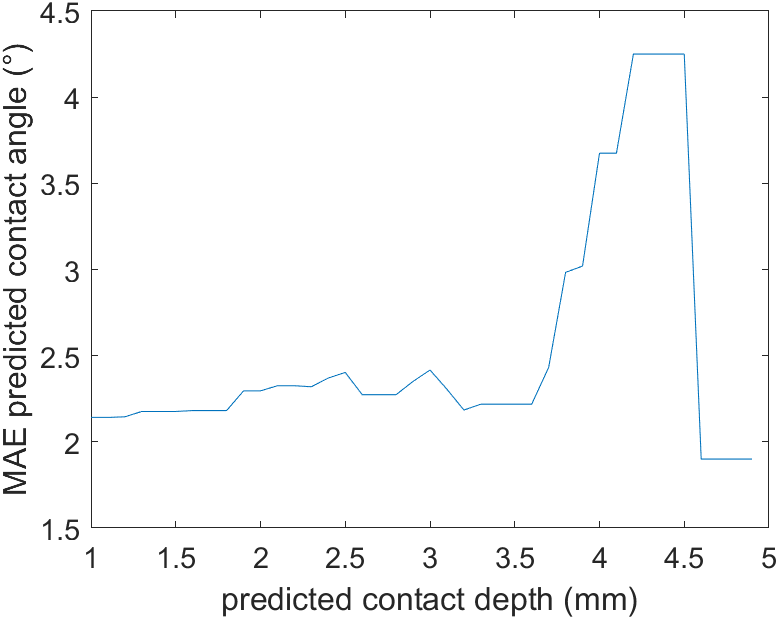}  \\ 	
	\end{tabular}
	\caption{PoseNet performance for predicted angle against (a) target relative angle and (b) predicted depth, collected using the TacMM system with the mobile manipulator.  }
	\label{fig:7a}
\end{figure}

In the online test with the TacMM system, the PoseNet model performance has angle errors mainly between 1° and 3° (Fig.~\ref{fig:7a}(a)), which as expected is less accurate than those with a robot arm. Likewise, the predicted angle errors are between 2° and 2.5° for predicted depths less than 3.6\,mm, after which there is a steep rise to over 4° (Fig.~\ref{fig:7a}(b)); there is also a discrepancy in the predictions of contact depth in this region on the robot arm (Fig.~\ref{fig:6}(a), top-right). Overall, we consider a good depth for making angle predictions to be 2.6\,mm, near the centre of the accurate predicted depth range and just above a region for poorer contact depth prediction performance of the PoseNet (Fig.~\ref{fig:6}(a)). We will use this as a threshold for the contact depth control strategy for the lifting task below.


The model operates in just 1\,ms, consistent with real time performance expectations for the pose-adjustment and object-lifting tasks.

\subsection{TacMM and Vision-based pose adjustment performances}
\label{sec:4b}

\begin{figure}[t!]
	\centering
    \begin{tabular}[b]{@{}c@{}c@{}}
           \textbf{\footnotesize{Simulated (a)}} & \textbf{\footnotesize{Simulated (b)}}  \\
		\includegraphics[width=0.49\columnwidth,trim={0 0 0 0},clip]{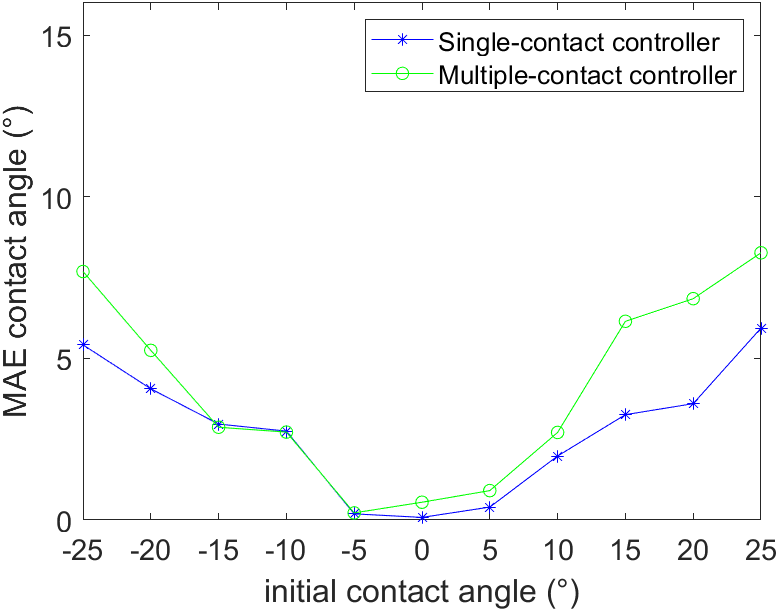} & 	
      	\includegraphics[width=0.49\columnwidth,trim={0 0 0 0},clip]{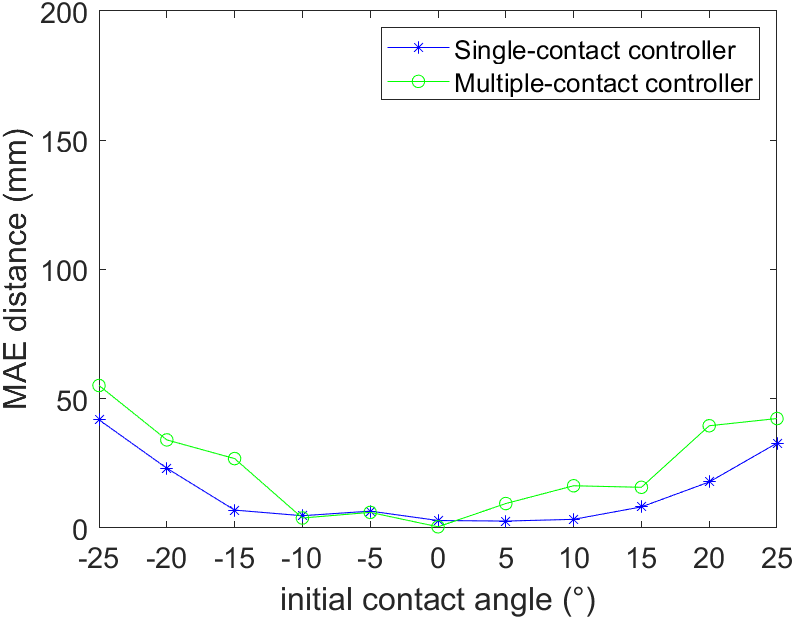} \\	
            \textbf{\footnotesize{TacMM (c)}} & \textbf{\footnotesize{TacMM (d)}} \\
		\includegraphics[width=0.49\columnwidth,trim={0 0 0 0},clip]{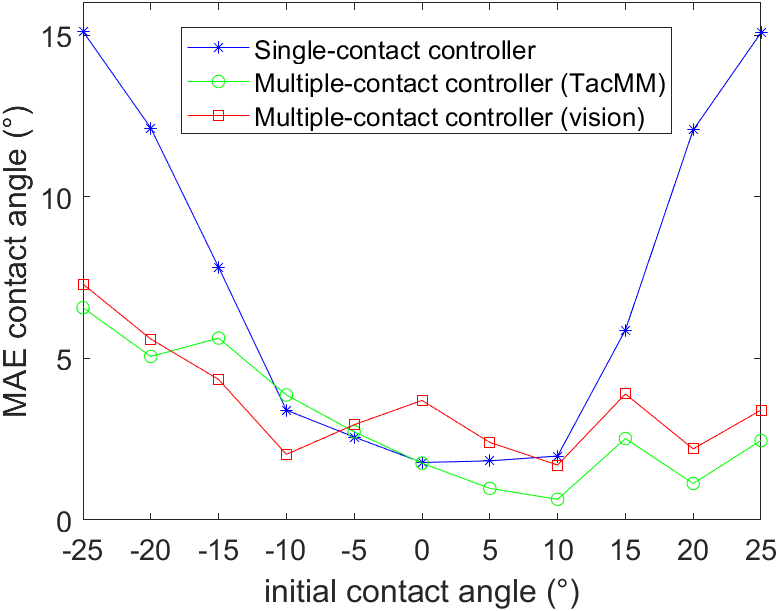} & 		
		\includegraphics[width=0.49\columnwidth,trim={0 0 0 0},clip]{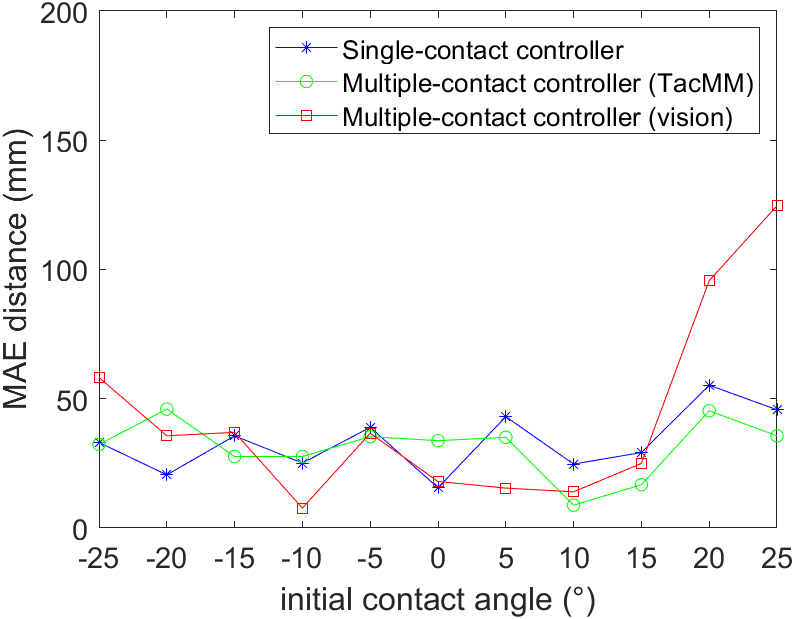} \\	
	\end{tabular}
	\caption{Performance of pose adjustment quantified by the MAE of the angle $\theta$ and translation distance $D$, in simulation for single- and multiple-contact strategies and on the TacMM system (with additional multiple contact strategy using vision).}
	\label{fig:7}
\end{figure}

\begin{figure*}[ht]
	\centering
	    {\bf (a) Load lifting under different attributes }\\
        \includegraphics[width=0.9\textwidth]{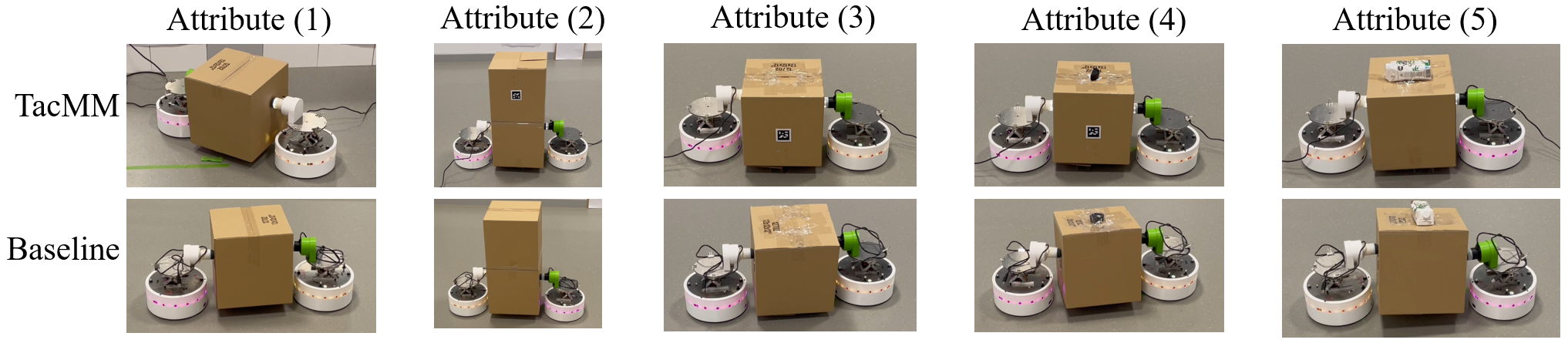} \\
	    {\bf (b) Failure cases relying on the purely vision-based baseline }\\
        \includegraphics[width=0.9\textwidth]{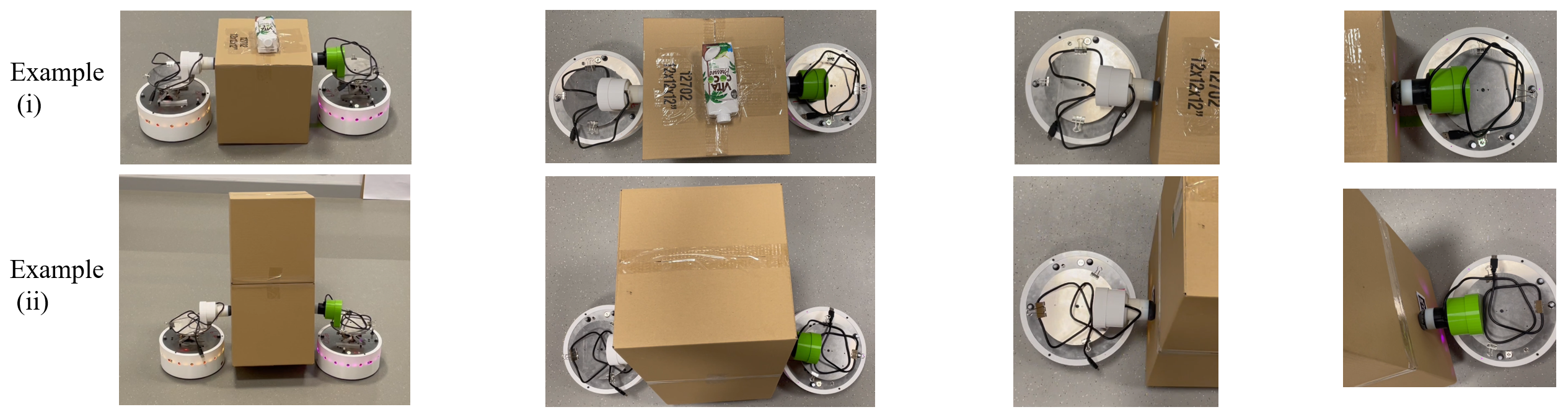}
	\caption{\textcolor{black}{Performance of load lifting with TacMM and the vision-based baseline. (a) Successful examples of load lifting with attributes including (1) an empty box, (2) two empty boxes stacked vertically, (3) a 200\,g weight on top of the box, (4) a 200\,g weight at the bottom of the box and (5) a 500\,g payload on top of the box. (b) Some failure cases when lifting using vision.}}
	\label{fig:8}
\end{figure*}


First, the pose adjustment strategy is tested in simulation to clarify the relationship between the errors and kinematics as the TacMM rotates around the object centre. The results show that both single- and multiple-contact strategies complete the task of placing the TacMM system against the object surface (Fig.~\ref{fig:7}(a,b)). For both strategies, the MAEs of the angle and distance increase with absolute initial angle to about 5° and 50\,mm errors, as expected because the larger overall movements introduce more errors. The single-contact controller seems slightly more accurate than the multiple-contact controller at large initial contact angles, which we attribute to the simulated TacMM pose estimation having low error, so the errors accumulate on multiple contacts.  


Second, the pose adjustment strategies are tested on the real mobile robot to examine how the real-world environment affects performance. The angle errors in the real-world are close to the sum of the error from the pose model and the robot kinematics (Fig.~\ref{fig:7}(c)). The single-contact TacMM controller performed poorly on angle, showing the benefits of a multiple-contact controller when there are prediction errors. The distance errors were similar for both single- and multiple-contact controllers (Fig.~\ref{fig:7}(d)). Overall, the multiple-contact TacMM system performed similarly to the multiple-contact vision-based system, except for poorer distance performance of the vision system at large initial contact angles. 

\begin{table}[t!]
    \centering
	\caption{Success rates of the TacMM and vision-based baseline in the lifting task.}
	\vspace{0em}
	\begin{tabular}{m{3cm}<{\centering}|m{1cm}<{\centering}  m{1cm}<{\centering}}
    	\textbf{Box attributes} & \textbf{TacMM} & \textbf{Baseline}\\
    	\hline\hline
    	Empty box & 80$\%$ & 50$\%$ \\
    	 \hline
    	Two empty boxes stacked vertically & 60$\%$ & 30$\%$ \\
    	 \hline
    	A 200g weight on the bottom of the empty box & 80$\%$ & 60$\%$ \\
    	 \hline
    	A 200g weight on the top of the empty box & 60$\%$ & 10$\%$ \\
    	 \hline
    	A 500g payload  on the top of the empty box & 50$\%$ & 10$\%$ \\
    	 \hline
    	Average successful rate & 66$\%$ & 32$\%$\\
    	 \hline
    	\end{tabular}
	\label{tab:2}
	\vspace{0em}
\end{table}



\subsection{Lifting task performance}

The success rate of each control system was compared for lifting various attributes of loads, with each load lifted ten times (Table~\ref{tab:2}). Although we saw in the previous section that the TacMM is similar to the purely vision-based (baseline) control systems for angle adjustment, we now see that the TacMM is clearly superior to the baseline in the load lifting task, with an average success rate improvement of 34$\%$ over all considered loads (Table~\ref{tab:2}, bottom line).

Depending on the attributes of the load, the success rate of the TacMM system can reach 40\% higher than the vision system as the payload weight increases (Table~\ref{tab:2}). This holds under demanding situations when the payload is close the maximum that the robots are able to lift, giving 50\% reliability for TacMM compared with only 10\% for with vision. Overall, TacMM outperforms the vision system in the load lifting task and shows promise to be improved further.

We believe the main reason for the TacMM's superior performance is its ability to accuracy predict and control the contact depth. This was not possible to verify in the previous section when comparing TacMM and vision-based pose adjustment because we did not have a independent measure of this depth at the required mm-scale, only of the angle and translation distance accuracy. 

In consequence of the superior depth performance, the deformations of the tactile sensors on the two robots are nearly the same, which enables two uniform forces to press on opposite sides of the load. In addition, the robots controlled by the TacMM system will only touch the box slightly instead of pushing the box. On the other hand, for the vision-based baseline, after the camera detects the pose of object according to the global view of the ArUco marker, controlling the contact depth relies just on the kinematics of the robot, which is less accurate. In consequence, the force and torques may not sum to near zero or the frictional force may not be enough to raise the object, resulting in more failures of lifting (Fig.~10).

\pagebreak 

\section{DISCUSSION}

{\color{black}Overall, this paper demonstrates the feasibility of TacMM tactile mobile manipulators for collaboratively performing load-lifting tasks with box objects typical of those in warehouses. Each TacMM comprises an open-sourced DOTS mobile robot equipped with a lifting platform, on which we mounted a 3D-printed open-source high-resolution tactile fingertip (TacTip) as an end effector. Thus, we extended a robot designed for researching automated logistics to use tactile sensing to interact with its surroundings.

A key aspect of lifting tasks with such robots is to adjust the pose of the end effector relative to the object to be lifted. For the pose adjustment stage, the TacMM system showed modest improvement over a purely vision-based system for the angle and translational position of the robot. However, for load lifting, the performance increase from touch became more pronounced because of the capability to control the force of the contact via adjusting the contact depth.

In this first study of tactile mobile manipulators we considered a minimal configuration of two TacMMs working collaboratively. Clearly, while two mobile manipulators can perform a basic task of lifting a regular box, they would be limited in handling more complex objects. Also, having just two small regions of contact will limit the robustness of the grasp, as the object could rotate or be more prone to slip. For this reason, we expect that TacMMs will be far more adept with three or more robots, and even be capable of sophisticated dexterity such as collaborative manipulation of the object. 

The challenge of scaling up to three or more TacMMs will involve developing control algorithms that use pose and other tactile information to collaboratively lift and handle complex objects. This has a close relation to the in-hand manipulation and grasping using multi-fingered robot hands with tactile fingertips. One could interpret the TacMMs as a reconfigurable robot hand where the fingertips can move around independently of each another. Thus, we hope that this study with two TacMMs could lead to a new cross-over between research on tactile dexterity with robot hands and the coordination of fleets of mobile robots for warehousing.


}


\bibliographystyle{IEEEtran}
\bibliography{RAL-format.bib}

\end{document}